%% file: main.tex
\documentclass[10pt,twocolumn,letterpaper]{article}

\usepackage{iccv}
\usepackage{times}
\usepackage{epsfig}
\usepackage{graphicx}
\usepackage{amsmath}
\usepackage{amssymb}
\usepackage{boldline}
\usepackage[dvipsnames]{xcolor}
\usepackage[normalem]{ulem}
\usepackage{enumitem}
\usepackage[accsupp]{axessibility}  % Improves PDF readability for those with disabilities.

 \usepackage{environ}
\newcommand{\acksection}{\section*{Acknowledgments}}
\NewEnviron{acks}{%
  \acksection
  \BODY
}

\usepackage[pagebackref=true,breaklinks=true,letterpaper=true,colorlinks,bookmarks=false]{hyperref}

\iccvfinalcopy % *** Uncomment this line for the final submission

\def\iccvPaperID{3944} % *** Enter the ICCV Paper ID here

\ificcvfinal\fi

\begin{document}

%%%%%%%%% TITLE
\title{Estimating Egocentric 3D Human Pose in Global Space}

\author{Jian Wang\textsuperscript{1,2}~~~~~Lingjie Liu\textsuperscript{1,2}~~~~~Weipeng Xu\textsuperscript{3}~~~~~Kripasindhu Sarkar\textsuperscript{1,2}~~~~~Christian Theobalt\textsuperscript{1,2}\\
\textsuperscript{1}MPI Informatics~~~~~\textsuperscript{2}Saarland Informatics Campus~~~~~\textsuperscript{3}Facebook Reality Labs\\
{\tt\small \{jianwang,lliu,ksarkar,theobalt\}@mpi-inf.mpg.de~~xuweipeng@fb.com}
}

% \author{Jian Wang\textsuperscript{1,2} \and Lingjie Liu\textsuperscript{1,2} \and Weipeng Xu\textsuperscript{3} \and Kripasindhu Sarkar\textsuperscript{1,2} \and Christian Theobalt\textsuperscript{1,2}}
% \author{
% \textsuperscript{1}MPI Informatics~~~~~\textsuperscript{2}Saarland Informatics Campus~~~~~\textsuperscript{3}Facebook Reality Labs}
% \author{Jian Wang \\
% MPI Informatics\\
% \and
% Lingjie Liu
% \and

\maketitle
\ificcvfinal\thispagestyle{empty}\fi

%%%%%%%%% ABSTRACT
\input{Sections/abstract}

%%%%%%%%% BODY TEXT

%-------------------------------------------------------------------------

% \JW{Todo: 
% R4: 1. Presentation needs to be clearer
% 2. Fair comparison with multi-frame based method (VIBE etc.)
% 3. Background is artificial: add real world result in introduction section of video
% }

% \JW{Todo: 
% R1: 1.  Segmentation of the body for SLAM (supp. mat.?)
% 2. Scale of SLAM for the Mo2Cap2 dataset. (supp. mat.?)
% 3. Computational requirements. (supp. mat.?)
% }

% \JW{Todo: 
% R3: 1. Details and comparisons of datasets. (supp. mat.?)
% 2. Details of sequential VAEs (supp. mat.?)
% }
\input{Sections/introduction}

\input{Sections/relatedwork}

\input{Sections/method}

\input{Sections/experiments}

\input{Sections/conclusions}

% \begin{acks}
% Jian Wang, Kripasindhu Sarkar and Christian Theobalt have  been supported by the ERC
% Consolidator Grant 4DReply (770784) and Lingjie Liu has been supported by Lise Meitner Postdoctoral Fellowship.
% \end{acks}

\noindent \textbf{Acknowledgments} Jian Wang, Kripasindhu Sarkar and Christian Theobalt have  been supported by the ERC Consolidator Grant 4DReply (770784) and Lingjie Liu has been supported by Lise Meitner Postdoctoral Fellowship.

{\small
\bibliographystyle{ieee_fullname}
\bibliography{egbib}
}

\end{document}

%% file: Sections/abstract.tex
\begin{abstract}
Egocentric 3D human pose estimation using a single fisheye camera has become popular recently as it allows capturing a wide range of daily activities in unconstrained environments, which is difficult for traditional outside-in motion capture with external cameras. However, existing methods have several limitations. A prominent problem is that the estimated poses lie in the local coordinate system of the fisheye camera, rather than in the world coordinate system, which is restrictive for many applications. Furthermore, these methods suffer from limited accuracy and temporal instability due to ambiguities caused by the monocular setup and the severe occlusion in a strongly distorted egocentric perspective. To tackle these limitations, we present a new method for egocentric global 3D body pose estimation using a single head-mounted fisheye camera. To achieve accurate and temporally stable global poses, a spatio-temporal optimization is performed over a sequence of frames by minimizing heatmap reprojection errors and enforcing local and global body motion priors learned from a mocap dataset. Experimental results show that our approach outperforms state-of-the-art methods both quantitatively and qualitatively.

\end{abstract}

%% file: Sections/introduction.tex
\section{Introduction}
% With a single downward looking fisheye camera, our goal is to estimate a person’s pose sequence for a variety of complex motions.

% Egocentric 3D human pose estimation is an important problem. 
%Traditional optical human motion capture techniques employ externally placed cameras that observe the human in an outside-in view. 
Traditional optical motion capture system with external, outside-in facing cameras is restrictive for many pose estimation applications that require the person to be able to roam around in a larger space, beyond a fixed recording volume. Examples are mobile interaction applications, pose estimation in large-scale workplace environments, or many AR/VR applications.
%With the high demand for capturing 3D human poses in an unconstrained environment, 
To enable this, methods for egocentric 3D human pose estimation using
% [WXU: let's be a bit more general here] 
% a single mobile camera (head- or body-mounted)
head- or body-mounted cameras were researched.
%Unlike traditional setups for 3D human pose estimation (e.g. a single or multiple cameras placed statically around the subject), 
These methods are mobile, flexible, and have the potential to capture a wide range of daily human activities even in large-scale cluttered environments.
%(e.g. cameras need to be close to the subject, no occluders in front of the subject, etc), 
%which makes capturing a wide range of human activities in daily life possible 
% for example for action recognition, performance analysis~\cite{DBLP:journals/tvcg/XuCZRFST19} and $x$R (including VR, AR and MR) applications~\cite{DBLP:conf/iccv/TomePAB19}. 

\begin{figure}
	\begin{center}
		\includegraphics[width=0.97\linewidth]{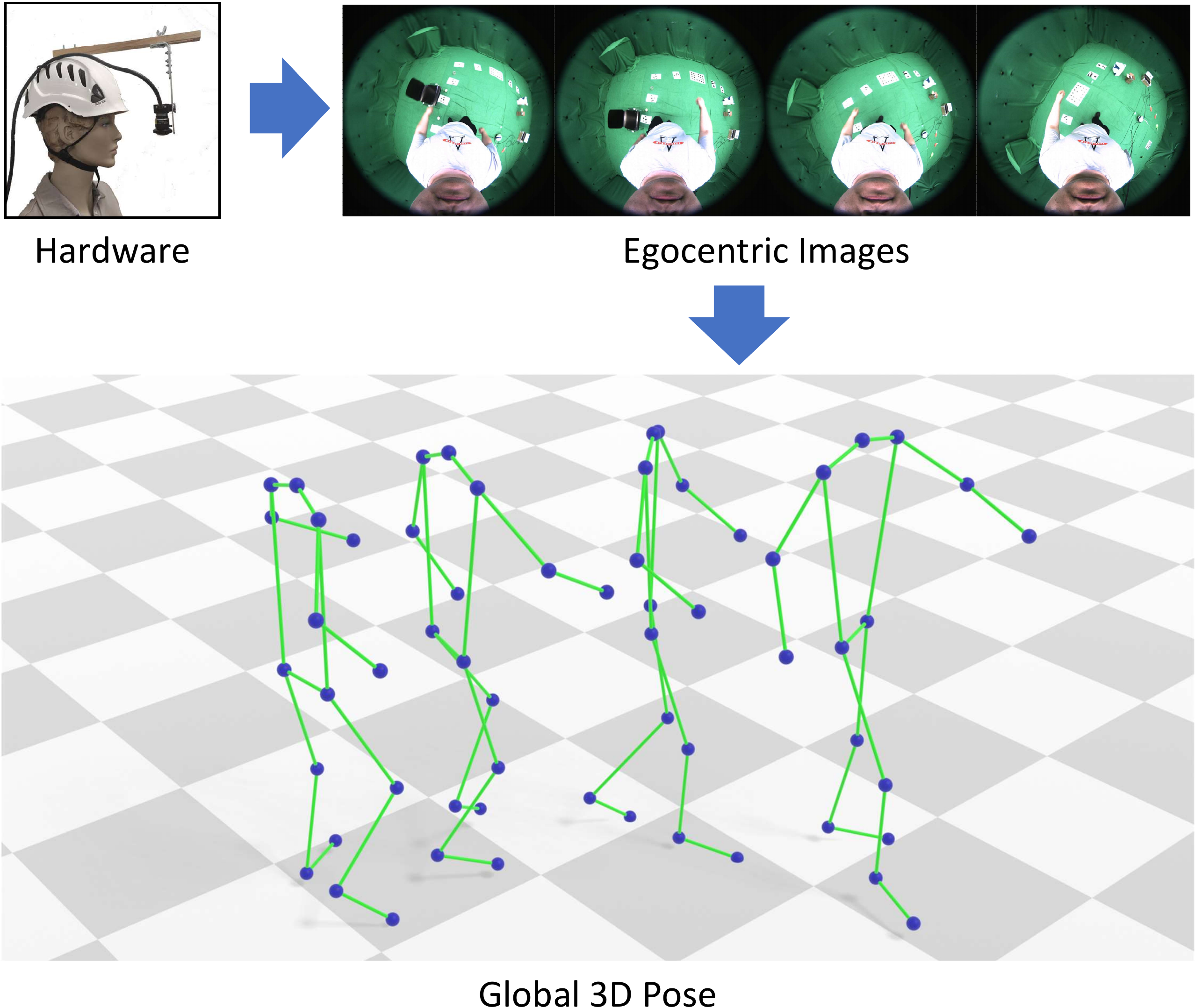}
	\end{center}
	\caption{Given challenging egocentric videos, our method produces realistic and accurate 3D global pose sequence.}
	\label{fig:io}
\end{figure}

%Limitations of existing works
Some egocentric capture methods study the estimation of face~\cite{Elgharib2020egocentricconferencing, elgharib2019egoface, DBLP:journals/tog/LiTOWTHNM15} and hand motions~\cite{Singh2016ActionRecon, SridharFastHandTracker, DBLP:conf/cvpr/MaFK16, DBLP:journals/pr/SinghAJ17}, while the estimation of the global full body pose has been less explored.  Mo$^2$Cap$^2$~\cite{DBLP:journals/tvcg/XuCZRFST19} and $x$R-egopose~\cite{DBLP:conf/iccv/TomePAB19} use a single head-mounted fisheye camera to capture the 3D skeletal body pose in a marker-less way. 
%the view of full body and estimate 3D body pose with respect to the fisheye camera. 
Both methods have demonstrated compelling 3D pose estimation results while still suffering from an important limitation: They estimate the local 3D body pose in egocentric camera space, while not being able to obtain the body pose with global position and orientation in the world coordinate system. Henceforth, we will refer to the former as ``local pose", in order to distinguish it from the ``global pose"  defined in the world coordinate system. Local pose capture alone is insufficient for many applications. For example, captured local body poses are not enough to animate the locomotion of a virtual avatar in $x$R environments, which requires global poses. 
% \CT{the following sentence may be not such a strong argument, maybe remove} Egocentric local pose capture can also cause ambiguities during action recognition; e.g, when disambiguating motions caused by turning the head or the entire body, for example.
% Obtaining the global translation and rotation of the body is difficult. One possible method is to simultaneously record the sequence with a multi-view setup. However,this would totally lose the advantages of using the mobile equipment. An alternative is to ... 

A straightforward solution is to simply project the local pose into the world coordinate system with the egocentric camera pose estimated by the SLAM.
However, the obtained global poses exhibit significant inaccuracies. First, they show notable temporal jitters as the video frames are processed independently without taking temporal frame coherence.  Second, they often show tracking failure due to the self-occlusion in the distorted view of the fisheye camera. 
%Third, the local body motion and the camera poses are inconsistent with each other because of errors in estimated local poses and camera poses.
% Third, the obtained global poses are less realistic because the local body motion and the camera poses are inconsistent with each other.
Third, the obtained global poses often show unrealistic motions (such as foot sliding and global jitters) due to the inconsistency between the local pose and the estimated camera pose, which are independent of each other.
% However, the obtained global poses often exhibit notable temporal jitters, because video frames are processed independently without taking temporal frame coherence into consideration. Furthermore, in the strongly distorted head-mounted fisheye perspective, the body is more often self-occluded, which may cause tracking failures. 

% \KS{However, the obtained global poses exhibit significant inaccuracies. Firstly, they show notable temporal jitters as the video frames are processed independently without taking temporal frame coherence. Secondly, they incorporate inaccurate global motion due to the unrobustness of the SLAM. Third, they often show tracking failure due to the self-occlusion in the distorted settings of the fisheye camera.}

%Introduce our method
To tackle these challenges, we propose a novel approach for accurate and temporally stable egocentric global 3D pose estimation with a single head-mounted fisheye camera, as illustrated in Fig.~\ref{fig:io}. 
In order to obtain temporally smooth pose sequences, we resort to a spatio-temporal optimization framework where we leverage the 2D and 3D keypoints from CNN detection as well as VAE-based motion priors learned from a large mocap dataset. 
The VAE-based motion priors have been proven effective to produce realistic and smooth motions in pose estimation methods like VIBE~\cite{DBLP:conf/cvpr/KocabasAB20} and MEVA\cite{Luo_2020_ACCV}. However, the RNN-based VAEs in these works are less efficient and unstable due to the vanishing and exploding gradients during our optimization process. Therefore, we propose a new convolutional VAE-based motion prior, which enables faster optimization speed and higher accuracy.
Furthermore, to reduce the error due to strong occlusion, we proposed a novel uncertainty-aware reprojection energy term by summing up the probability values at the pixels on the heatmap occupied by the projection of the 3D estimated joints rather than comparing the projection of 3D estimated joints against the predicted 2D joint position. 
Finally, in order to make the local body poses consistent with the camera poses estimated by SLAM, we introduce a global pose optimizer with a separate VAE.

We evaluate our method on the dataset provided by Mo$^2$Cap$^2$~\cite{DBLP:journals/tvcg/XuCZRFST19} and also a new benchmark we collected with 2 subjects performing various motions. 
% With the publication we will make our dataset publically available. \KS{Maybe write 1 line about the dataset.}
Our method outperforms the state-of-the-art methods both quantitatively and qualitatively. Our ablative analysis confirms the efficacy of our proposed optimization algorithm with learned motion prior and uncertainty-aware reprojection loss for improved local and global accuracy and temporal stability. 
To summarize, our technical contributions are as follows:

% \WX{Say something stronger to emphasize the vae pose prior.
% For instance, the motion priors help to reduce the depth ambiguity and the uncertainty of the detection due to occlusions.
% }
%Our extensive ablation studies demonstrate that incorporating motion priors into egocentric body pose estimation efficiently improves both the local pose and global pose in terms of accuracy and temporal stability, and that our new uncertainty-aware reprojection loss further increases the accuracy of local pose estimation so leads to better global pose estimation. 

%%%%%%%%Old contributions
%In summary, our technical contributions are: First, we developed a novel framework for accurate and temporally stable global 3D human pose estimation from a monocular egocentric video. 
%Second, we designed a optimization algorithm with the assistance of local and global motion prior captured by a convolutional network based VAE.
%Third, an uncertainty-aware reprojection loss is employed to alleviate the influence of self-occlusions in the captured video. Fourth, the body movement is constrained to be consistent with the locomotion with a global pose optimizer.
%%%%%%%%%%%

% \vspace{-0.2cm}
\begin{itemize}[leftmargin=*]
\itemsep 0.0em
\item A novel framework for accurate and temporally stable global 3D human pose estimation from a monocular egocentric video. 
% \item A novel framework for estimating accurate and temporally stable 3D human pose in global space from a monocular egocentric video. 
% \item A new CNN based VAE to capture the pose prior and an optimization algorithm that is more accurate and efficient than the previously used RNN based VAE.
\item A new optimization algorithm with the assistance of local and global motion prior captured by an efficient convolutional network based VAE.
\item An uncertainty-aware reprojection loss to alleviate the influence of self-occlusions in the egocentric settings. 
\item Our method outperforms various baselines in terms of the accuracy of the estimated global and local pose. 
\end{itemize}
% \vspace{-0.2cm}

Our method works for a wide range of motions in various environments. We recommend watching the video in \url{http://gvv.mpi-inf.mpg.de/projects/globalegomocap} for better visualization.

% \KS{Shall we claim that we are the first method for global egocentric pose estimation?}
% \begin{itemize}
%     \item \JW{}
%     \item \JW{A combination of CNN-based local pose and optimization-based global pose estimation with the assistance of separate local and global motion priors.}
%     \item A new uncertainty-aware reprojection loss to alleviate the adverse influence of the perspective distortion and self-occlusions in the captured video.
% \end{itemize}

% \begin{itemize}
%     \item A novel approach for global 3D human pose estimation from an egocentric fisheye camera.
%     \item A combination of CNN-based local pose and optimization-based global pose estimation with the assistance of separate local and global motion priors.
%     \item A new uncertainty-aware reprojection loss to alleviate the adverse influence of the perspective distortion and self-occlusions in the captured video.
% \end{itemize}

%% file: Sections/relatedwork.tex
\section{Related Work}

\paragraph{Egocentric 3D full body pose estimation}
Capturing full-body motion from an egocentric camera perspective has attracted more and more attention in recent years while it is challenging as it is difficult to observe the whole body from close proximity in the egocentric setting. 
% Capturing full body motion from an egocentric camera perspective has been widely explored in recent years. 
% For example, some methods estimate facial expression or eye movements with a head-mounted camera or a VR heatset  \cite{Elgharib2020egocentricconferencing,elgharib2019egoface,thies16FaceVR, Lombardi2018DeepFace, cha2018fullycapture, DBLP:conf/uist/SuganoB15, DBLP:journals/tog/LiTOWTHNM15}, while some other methods estimate user's gesture and activity  by using a head-mounted or chest-worn camera \cite{Singh2016ActionRecon,SridharFastHandTracker,DBLP:conf/cvpr/MaFK16,DBLP:conf/iccv/CaoZ0LC17,OccludedHands_ICCV2017,DBLP:conf/cvpr/OhnishiKKH16, DBLP:journals/pr/SinghAJ17}. However, they only focus on the motion estimation of body parts, as it is difficult to observe the whole body from a close proximity in the egocentric setting. 
% To solve this, some works adopt the inside-out configuration. 
Some works estimate the full-body pose by analyzing the motion of the observed environment. Shiratori \etal \cite{DBLP:journals/tog/ShiratoriPSSH11} attach 16 cameras to the subject’s limbs and torso to recover the human pose by performing SFM of the environment. Jiang and Grauman \cite{DBLP:conf/cvpr/JiangG17a} reconstruct full-body pose by leveraging learned dynamic and pose coupling over a long time span. Yuan and Kitani \cite{yuan20183d,DBLP:conf/iccv/YuanK19} use video-conditioned control techniques to estimate and forecast physically-plausible human body motion.
%However, it is impossible to get the exact body pose without any information of the human body itself. 
%
% reducing for space
%Although their results are physically plausible, they exhibit inaccuracy often due to the lack of direct observation of the human body.
%it is impossible to get the exact body pose without any information of the human body itself. 
%
% reducing for space
%Another way of addressing the problem of field-of-view in the egocentric setting is the use of multiple cameras.
%Some other papers mount wide-view fisheye cameras on the head, enabling large field of view which most of the body is in.  
Rhodin \etal~\cite{DBLP:journals/tog/RhodinRCISSST16} are the first to propose a full-body capture method with a helmet-mounted stereo fisheye camera. Cha \etal~\cite{cha2018fullycapture} estimate the 3d body pose from two head-mounted pinhole cameras with a recurrent neural network.
To avoid inconvenience of large setup, some researchers use a single wide-view fisheye camera. %
Xu \etal~\cite{DBLP:journals/tvcg/XuCZRFST19} and Tome \etal~\cite{DBLP:conf/iccv/TomePAB19} use a compact monocular setting and developed learning-based approaches to estimate ego-pose from a single frame. Hwang \etal \cite{Hwang2020} mount an ultra-wide fisheye camera on the user's chest and estimate body pose, camera rotation and head pose from a single fisheye image. 
However, these methods neither exploit temporal consistency, nor ensure the reality of predicted motion. Our method, on the contrary, leverages motion prior based optimization approach to make the prediction consistent and accurate. 
% All these methods have demonstrated compelling 3D pose estimation results, but still have several limitations. First, they estimate the local 3D body pose in egocentric camera space, but not global rotation and translation in 3D world coordinates. Second, these methods do not exploit temporal consistency and show notable temporal jitter. Our method, on the contrary, estimate global 3D human body pose which is stable, temporally coherent and more accurate.

\begin{figure*}[h]
	\begin{center}
		\includegraphics[width=0.97\linewidth]{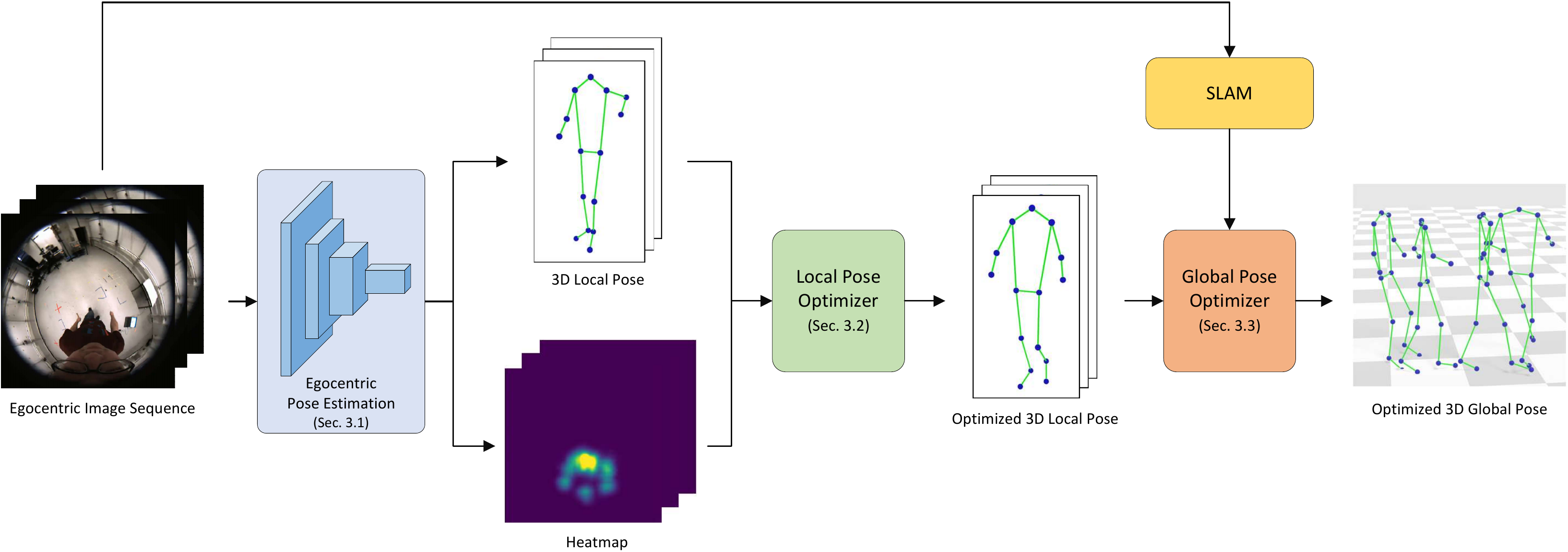}
	\end{center}
	\caption{Overview of our method. Our method takes an egocentric video as input and processes it in segments. For each segment consisting of a fixed number of consecutive frames, we first apply an egocentric pose estimation method to obtain initial 3D local poses and 2D  heatmaps which are then fed into the local pose optimization framework to get optimized local poses. Next, combined with the camera poses estimated from ORB-SLAM2, the optimized 3D local poses are transformed from the local egocentric camera space to the world coordinate space and then optimized via the global pose optimization to produce the final global poses. 
% 	first use some available egocentric 3D pose estimation method to predict the 3D pose and 2D heatmap from each frame in the input sequence. Then the 3D pose and heatmap is fed into our local pose optimizer to get the optimized local pose sequence. Finally, combined with the camera pose from the SLAM, we can get the optimized global pose from our global pose optimizer.
	}
	\label{fig:framework}
\end{figure*}

\paragraph{Leveraging learned prior in 3D pose estimation}

In order to enhance the accuracy of pose estimation and make predictions more realistic, a lot of recent methods leverage the prior learned from the mocap dataset. 
%in the 3D pose estimation task. 
Some of them capture the prior in the Gaussian space. For example, Bogo \etal \cite{DBLP:conf/eccv/BogoKLG0B16} and Arnab \etal \cite{DBLP:conf/cvpr/ArnabDZ19} captures the prior to optimize the SMPL body model \cite{SMPL:2015} by fitting a mixture of Gaussians to CMU mocap dataset \cite{CMUmocap}. Pavlakos \etal \cite{DBLP:conf/cvpr/PavlakosCGBOTB19} train a VAE to learn priors of SMPL parameters on AMASS dataset, which contains richer varieties of human motions. Zanfir \etal\cite{zanfir2020weakly} use normalizing flow in order to avoid the compromise between KL divergence and reconstruction loss in VAE. 
Some other methods incorporate the pose prior by training a generative adversarial network (GAN). Yang \etal \cite{DBLP:conf/cvpr/YangO0RLW18} develop a adversarial learning framework with multi-source discriminator. Kanazawa \etal \cite{DBLP:conf/cvpr/KanazawaBJM18, DBLP:conf/cvpr/KanazawaZFM19} and Zhang \etal \cite{DBLP:conf/iccv/ZhangFKM19} train discriminators for each joint rotation parameter to tell if these parameters are realistic.
% come from a real human shape and pose.
Kocabas \etal \cite{DBLP:conf/cvpr/KocabasAB20} propose a temporal network architecture with an RNN-based discriminator for the adversarial training on the sequence of SMPL parameters.
Different from previous methods, our method captures the global motion prior learned with a light-weight sequential VAE, which enables direct optimization in the global coordinate system.

% \vspace{-0.5em}
\paragraph{Monocular 3D pose estimation in video}
% Monocular 3D pose estimation has been the focus of research for a long time while a lot of methods process single images and therefore exhibit notable temporal jitter in a video sequence.
Monocular 3D pose estimation has been the focus of research for a long time. Some methods predict 2D joints and perform 2D-to-3D lifting separately \cite{DBLP:conf/cvpr/ChenR17, DBLP:conf/iccvw/JahangiriY17,DBLP:conf/iccv/MartinezHRL17}, while some other methods regress the 3D pose directly \cite{DBLP:conf/accv/LiC14, DBLP:conf/3dim/MehtaRCFSXT17,DBLP:conf/bmvc/TekinKSLF16,DBLP:journals/corr/TekinMSF16,DBLP:conf/iccv/KolotourosPBD19}. These methods process single image and therefore exhibit notable temporal jitter in a video sequence.
To solve this, many recent methods exploit temporal information from the video.
%Recently, there have been more efforts exploiting temporal information from video to reduce the influence of noise and increase prediction accuracy. 
Zhou \etal \cite{DBLP:conf/cvpr/ZhouZLDD16} introduce EM method to estimate 3D pose from 2D predictions over the entire sequence. Mehta \etal \cite{VNect_SIGGRAPH2017} and Du \etal \cite{DBLP:conf/eccv/DuWLHGWKG16} apply temporal filtering across 2D and 3D poses.
% to predict a temporally consistent 3D pose. 
Lin \etal \cite{DBLP:conf/cvpr/LinLLWC17}, Hossain \etal \cite{DBLP:conf/eccv/HossainL18}, Kocabas \etal \cite{DBLP:conf/cvpr/KocabasAB20} and Katircioglu \etal \cite{DBLP:journals/ijcv/KatirciogluTSLF18} use recurrent networks to predict 3D pose sequences by leveraging previously predicted 2D and 3D poses. Pavllo \etal \cite{DBLP:conf/cvpr/PavlloFGA19} generates 3D poses with temporal-convolution, while Cai \etal \cite{DBLP:conf/iccv/CaiGLCCYM19} and Wang \etal \cite{wang2020motion} leverage graph convolutional network to capture the temploral information. Luo \etal \cite{luo20203d} firstly get coarse motion with a GRU based human motion VAE and then refine the motion with a residual estimation network. Different from all previous works, our method capture the motion prior with a 1D convolution based sequential VAE, and we use the VAE in our optimization framework.
%Different from all previous works, our method capture the temporal context with the 1D convolutional VAE, and we use the captured context as motion prior to optimize the 3D pose sequence. 

%% file: Sections/method.tex
\section{Method}

Our goal is to estimate the global body poses from a video sequence captured by a head-mounted fisheye camera.
We provide an overview of our pipeline in Fig.~\ref{fig:framework}. 
The video frames are split into segments with $B$ frames each ($B=10$ in our experiments).
Our pipeline takes one segment consisting of $B$ consecutive frames, $\mathcal{I}_{seq}=\{\mathcal{I}_1,\dots,\mathcal{I}_B\}$, as inputs and outputs the global poses of all the individual frames, $\mathcal{P}_{seq}^g=\{\mathcal{P}_1^g,\dots,\mathcal{P}_B^g\}$. 
% Note that each video frame may receive several candidate global poses and the final pose is computed by averaging all of its candidate poses. 
For each segment, we first calculate the 3D local pose and 2D heatmap of each frame using an egocentric local body pose estimation method (Sec.~\ref{sec:pose_estimation}). 
Next, we learn the local motion prior from local motion sequences of the AMASS dataset~\cite{AMASS:ICCV:2019} with a sequential VAE~\cite{DBLP:journals/corr/KingmaW13} (Sec.~\ref{subsec:vae}), and perform a spatio-temporal optimization with the local motion prior by minimizing the heatmap reprojection term and several regularization terms (Sec.~\ref{subsec:local_optim}). 
% Note that in order to consider the uncertainty existed in the 2D joint detection in the reprojection term, we calculate the reprojection term in a new way based on 2D heatmap . 
Given the optimized local poses, we transform them from local fisheye camera space to the world coordinate system with camera poses estimated by a SLAM method to get initial global poses (Sec.~\ref{subsec:slam}). 
To improve global poses, we learn the global pose prior by training a second sequential VAE on the global motion sequences of the AMASS dataset, and impose the global pose prior in a spatio-temporal global pose optimization (Sec.~\ref{subsec:global_optim}). 
Please refer to the supplementary materials for our implementation details.

\subsection{Local Pose Estimation}\label{sec:pose_estimation}
% In a preprocessing stage, g
Given a segment containing $B$ consecutive frames $\mathcal{I}_{seq}$,
% $\mathcal{I}_{seq}=\{\mathcal{I}_1,\dots,\mathcal{I}_B\}$
we estimate local poses represented by 15 joint locations $\widetilde{\mathcal{P}}_{seq}=\{\widetilde{\mathcal{P}}_1,\dots,\widetilde{\mathcal{P}}_B \}$, $\widetilde{\mathcal{P}}_i \in \mathbb{R}^{15 \times 3}$, and 2D heatmaps $\mathcal{H}_{seq}=\{\mathcal{H}_1,\dots,\mathcal{H}_B \}$ using an egocentric local pose estimation method. 
Note that our approach can work with any egocentric local pose estimation methods. In our experiments, we evaluate our approach on the results of two state-of-the-art methods: Mo$^2$Cap$^2$ \cite{DBLP:journals/tvcg/XuCZRFST19} and $x$R-egopose \cite{DBLP:conf/iccv/TomePAB19}. 

% To achieve this, we can use any single-frame based egocentric pose estimation method as long as it meets the input/output requirement. In our experiment, we use two available methods: the Mo$^2$Cap$^2$ \cite{DBLP:journals/tvcg/XuCZRFST19} and the $x$R-egopose \cite{DBLP:conf/iccv/TomePAB19}. 

% The Mo$^2$Cap$^2$ trains a 2D heatmap prediction network on the synthetic dataset, then uses the features from 2D network to train the network to predict the distance of each joint. The actual joint positions are recovered by projecting the 2D joint detections using the distance estimation and the fisheye camera model. This method guarantees the 3D pose can be perfectly re-projected to detected 2D joints but also easily ends up with unrealistic poses and bone lengths. Different from the Mo$^2$Cap$^2$, the $x$R-egopose train a autoencoder to directly generate 3D pose from 2d heatmaps. This method can usually avoid unrealistic poses, but still suffers from the loss of re-projection accuracy.

% It will be shown in our experiment section that our optimization approach works well for both of the aforementioned methods, which proves the remarkable versatility of our framework.

\subsection{Local Pose Optimization}\label{sec:local_pose}
Although Mo$^2$Cap$^2$ and $x$R-egopose can produce compelling results, both approaches suffer from limited accuracy and temporal instability, which is mainly due to depth ambiguities caused by the monocular setup and severe occlusions in a strongly distorted egocentric perspective. 
To improve local poses, we design an efficient spatio-temporal optimization framework which first learns the local pose prior as a latent space 
%from local motion sequences of the large-scale AMASS dataset~\cite{AMASS:ICCV:2019} 
with a sequential VAE~\cite{DBLP:journals/corr/KingmaW13} (Sec.~\ref{subsec:vae}) and then searches for a latent vector in the learned latent space by minimizing a reprojection term and some regularization terms (Sec.~\ref{subsec:local_optim}). 

% In this section, we formulate an local motion prior based optimization framework to generate the realistic and accurate local pose sequence from the output of Sec.~\ref{sec:pose_estimation}. We firstly train a sequential VAE to capture the local motion prior in the AMASS dataset (Sec.~\ref{subsec:vae}), then we minimize the energy function by considering the re-projection of 3D keypoints, temporal consistency of joint positions, and bone length (Sec.~\ref{subsec:optim}). 

% \subsubsection{Sequential VAE}\label{subsec:vae}
\subsubsection{Learning Motion Prior}\label{subsec:vae}
% \begin{figure}
% 	\begin{center}
% 		\includegraphics[width=0.97\linewidth]{Figures/VAE/vae-imgs-crop.pdf}
% 	\end{center}
% 	\caption{Architecture of sequential VAE.}
% 	\label{fig:vae}
% \end{figure}

% To construct a latent space encoding local motion prior, we train a sequential VAE~\cite{DBLP:journals/corr/KingmaW13} on local motion sequences of the AMASS dataset~\cite{AMASS:ICCV:2019} which are split into batches with batch size $B$ for training. 
To construct a latent space encoding local motion prior, we train a sequential VAE~\cite{DBLP:journals/corr/KingmaW13} on local motion sequences of the AMASS dataset~\cite{AMASS:ICCV:2019} which are split into segments for training. 
We denote a segment consisting of $B$ consecutive poses as $\mathcal{Q}_{seq}=\{\mathcal{Q}_1, \dots, \mathcal{Q}_B \} (\mathcal{Q}_i \in \mathbb{R}^{15 \times 3})$.
% which is obtained by putting a virtual fisheye camera attached to the forehead of the human mesh at a distance similar to our real-world settings.
The sequential VAE consists of an encoder $f_{enc}$ and a decoder $f_{dec}$.
The encoder is used to map an input sequence of human local poses $\mathcal{Q}_{seq}$ to a latent vector $z$, and the decoder is used to reconstruct a pose sequence,
$\widehat{\mathcal{Q}}_{seq}=\{\widehat{\mathcal{Q}}_1, \dots, \widehat{\mathcal{Q}}_B \} (\widehat{\mathcal{Q}}_i \in \mathbb{R}^{15 \times 3})$, from the latent vector. 
Following \cite{DBLP:journals/corr/KingmaW13}, the training loss of VAE is formulated as:
% Same as \cite{DBLP:journals/corr/KingmaW13}, the encoder and decoder are trained by minimizing the reconstruction loss between the input pose sequence and the decoded pose sequence and a regularization term which keeps the latent space close to a normal distribution. 
% The training loss is formulated as follows:
% By using a 5-layer 1D convolutional encoder $f_{encoder}$ and a 5-layer 1D convolutional decoder $f_{decoder}$, our sequential VAE extracts the latent features $z$ from sequence of joint positions $P_{seq} = \{P_1,\dots, P_B\}$ and reconstructs the input sequence from the latent space. According to  \cite{DBLP:journals/corr/KingmaW13}, the training loss of the VAE is formulated as:
\begin{equation}
\begin{aligned}
    \mathcal{L}_{total} &= c_1\left\Vert \widehat{\mathcal{Q}}_{seq} - \mathcal{Q}_{seq} \right\Vert_2^2\\
    &+ c_2 KL[q(z\vert \mathcal{Q}_{seq}) \Vert \mathcal{N}(0, I)]
\end{aligned}
\end{equation}
where $z = f_{enc}(\mathcal{Q}_{seq})$, $\widehat{\mathcal{Q}}_{seq} = f_{dec}(z)$, $q(z\vert \mathcal{Q}_{seq})$ refers to the projected distribution of $\mathcal{Q}_{seq}$ in the latent space, $\mathcal{N}(0, I)$ refers to the standard normal distribution, and $KL(.)$ refers to the Kullback–Leibler divergence.

% \JW{The RNN-based VAEs have been proven efficient either as a loss term for training network \cite{DBLP:conf/cvpr/KocabasAB20}, or as a network component for obtaining rough estimation \cite{Luo_2020_ACCV}. However, our optimization framework}

Different from previous pose estimation methods \cite{DBLP:conf/cvpr/KocabasAB20, Luo_2020_ACCV} which leverage RNN-based VAEs to capture the motion prior, both the encoder $f_{enc}$ and the decoder $f_{dec}$ of our sequential VAE are designed as 5-layer 1D convolutional networks. Comparing with RNN-based VAEs, the convolutional networks in our sequential VAE is more efficient in the optimization iterations since it can be parallelized over time sequence. Moreover, the RNNs suffer from vanishing and exploding gradients more easily, which makes optimization process less stable. We have compared the sequential VAE in our method with RNN-based VAEs in VIBE~\cite{DBLP:conf/cvpr/KocabasAB20} and MEVA~\cite{Luo_2020_ACCV} in Sec.~\ref{sec:ablation}. More details of sequential VAE is shown in the supplementary materials.

\subsubsection{Optimizing Local Poses with Local Motion Prior}\label{subsec:local_optim}
With the learned latent space of local motion, the task of optimizing local poses with the local motion prior can be formulated as the problem of finding a latent vector $z$ in the learned latent space such that the reconstructed local pose sequence $\mathcal{P}_{seq} = f_{dec}(z)$ minimizes the following objective function:

% With the trained sequential VAE, the task of optimization method is to find a latent vector $z$ such that we can get the local pose sequence $P_{seq} = f_{decoder}(z)$ which minimizes the objective function $E (P_{seq})$. The overall objective function is shown as follow:
\begin{equation}
\begin{aligned}
	E (\mathcal{P}_{seq}) &= \lambda_R E_{R}(\mathcal{P}_{seq}) + \lambda_J E_{J}(\mathcal{P}_{seq}, \widetilde{\mathcal{P}}_{seq}) \\
	&+ \lambda_T E_{T}(\mathcal{P}_{seq}) + \lambda_B E_{B}(\mathcal{P}_{seq}) \label{eq:optim}
\end{aligned}
\end{equation}
where $E_{R}(.), E_{J}(.), E_{T}(.), E_{B}(.)$ are the reprojection term, pose regularization term, motion smoothness regularization term and bone length regularization term, respectively, which we will describe in detail later. In our experiment, we set the weights $\lambda_R = 0.01$, $\lambda_J = 0.01$, $\lambda_T = 1$ and $\lambda_B = 0.01$, respectively.

\paragraph{Heatmap-based Reprojection:} 

Previous works \cite{DBLP:conf/cvpr/ArnabDZ19, DBLP:conf/eccv/BogoKLG0B16, DBLP:conf/cvpr/PavlakosCGBOTB19,DBLP:journals/corr/abs-2003-10350} calculate the reprojection term by summing up the Euclidean distance values between the projection of estimated 3D joints and detected 2D joints. 
However, this calculation is sensitive to 2D joint detection errors due to the strong self-occlusions caused by the egocentric perspective.
To tackle this issue, we define a heatmap-based reprojection error by leveraging the uncertainty captured in the predicted 2D heatmaps, where the value at each pixel describes the probability of this pixel being a 2D joint. 
This new reprojection term is calculated by maximizing the summed heatmap values at the reprojected 2D joint positions:
% Instead, we define the reprojection term based on 2D heatmap where the value at each pixel describes the probability of this pixel being a 2D joint. The definition of heatmap-based reprojection term is defined as: 

% Error will be introduced if the 2D detections are not accurate, while in the egocentric camera viewpoint, the extreme distortion and serious self-occlusions are making the accurate detection of 2D pose quite difficult. We find that the predicted heatmap is able to capture the uncertainty in the 2D pose estimation, which is also confirmed by Tome et al. \cite{DBLP:conf/iccv/TomePAB19}. Based on this observation, we design our heatmap reprojection error (Eq.~\ref{eq:reproj}) by maximizing the heatmap values at the reprojected 2D joint positions. Since the heatmap represents the probability of possible 2D joint positions, this error function is also maximizing the overall probability of reprojected joints. The formulation of heatmap reprojection error is shown as follow:

% \begin{equation}
% 	E_{R}(P_{seq}) = -\sum_{P_i\in P_{seq}}\left\Vert\text{HM}_i(\Pi(P_{i}))\right\Vert_2^2 \label{eq:reproj}
% \end{equation}

\begin{equation}
	E_{R}(\mathcal{P}_{seq}) = -\sum_{i=1}^{B}\left\Vert\text{HM}_i(\Pi(\mathcal{P}_{i}))\right\Vert_2^2 \label{eq:reproj}
\end{equation}
where $\text{HM}_i(.)$ returns the value at a pixel on $\mathcal{H}_i$, the heatmap of $i$-th frame. $\Pi(.)$ refers to the projection of a 3D point. 
Specifically, the projection of a 3D point $[x, y, z]^T$ can be written as:
% where $\text{HM}_i$ is the heatmap of $i$-th frame and $\text{HM}_i(\textbf{x})$ returns the heatmap value at 2D position  $\textbf{x}$. $\Pi$ is the reprojection function based on the fisheye camera model \cite{scaramuzza2014omnidirectional}. Given the calibration of the fisheye camera, each 2D joint position $[u,v]^T$ can be mapped from its corresponding 3D ray vector $[x, y, z]^T$ with respect to the fisheye camera coordinate system:

\begin{equation}
	[u, v]^T = \frac{[x, y]^T}{\sqrt{x^2 + y^2}} \times f(\rho)
\end{equation}
where $\rho = \arctan(z / \sqrt{x^2 + y^2})$ and $f(\rho) = \alpha_0 + \alpha_1 \rho + \alpha_2 \rho^2 + \alpha_3 \rho^3 +\dots$ is a polynomial  obtained from camera calibration.

\paragraph{Pose Regularization:} 
To constrain the optimized pose $\mathcal{P}_{i}$ to stay close to the initial pose $\widetilde{\mathcal{P}}_{i}$, we define the pose regularizer as:

% We use this term (Eq.~\ref{eq:init}) to encourage each pose $P_{i}$ in the solution sequence $P_{seq}$ to stay close to our initial pose estimation $\widetilde{P}_{i}$ in initial sequence $\widetilde{P}_{seq}$ by penalizing the difference between the $P_{i}$ and $\widetilde{P}_{i}$:

\begin{equation}
	E_{J}(\mathcal{P}_{seq}, \widetilde{\mathcal{P}}_{seq}) = \sum_{i=1}^{B}\left\Vert \mathcal{P}_{i} - \widetilde{\mathcal{P}}_{i} \right\Vert_2^2 \label{eq:init}
\end{equation}

\paragraph{Motion Smoothness Regularization:} 
% Following \cite{XNect_SIGGRAPH2020}, the temporal smoothness error (Eq.~\ref{eq:smooth}) encourages smooth motions that are typical of humans in videos. This is achieved by penalizing the acceleration of each joint over the whole sequence:
Same as \cite{XNect_SIGGRAPH2020}, the temporal smoothness regularizer (Eq.~\ref{eq:smooth}) is used to improve the temporal stability of the estimated poses, which is calculated based on the acceleration of each joint over the whole sequence:
\begin{equation}
	E_{T}(\mathcal{P}_{seq}) = \sum_{i=2}^{B}\left\Vert \nabla \mathcal{P}_{i} - \nabla \mathcal{P}_{i-1}   \right\Vert_2^2 \label{eq:smooth}
\end{equation}
where $\nabla \mathcal{P}_i = \mathcal{P}_i - \mathcal{P}_{i-1}$.
% where the rate of change in joint positions, $\nabla P$, is approximated by using backward differences.
\paragraph{Bone Length Regularization:}
To explicitly enforce the constraint that each bone length stay fixed, we define the bone length regularizer as the difference between the bone length and the average bone length over the pose sequence. 

% The bone length regularizer (Eq.~\ref{eq:bone}) penalizes the difference between joint position $P_i$ and the average bone length over the pose sequence. In this way, we can encourage the bone length of each pose $P_i$ in the sequence $P_{seq}$ to be consistent.
% \begin{equation}
% 	E_{B}(P_{seq}) = \sum_{P_i\in P_{seq}}\left\Vert L_{P_{i}} - \frac{1}{B} \sum_{P_j\in P_{seq}}L_{P_{j}}\right\Vert_2^2 \label{eq:bone}
% \end{equation}
\begin{equation}
	E_{B}(\mathcal{P}_{seq}) = \sum_{i=1}^{B}\left\Vert L_{\mathcal{P}_{i}} - \frac{1}{B} \sum_{j=1}^{B}L_{\mathcal{P}_{j}}\right\Vert_2^2 \label{eq:bone}
\end{equation}
where the $ L_{\mathcal{P}_{i}}$ is a vector composed of the length of each bone of 3D pose $\mathcal{P}_i$. 

% \LJ{Move to 'Experiments'}
% Here we note that different from previous optimization systems based on priors captured by VAE \cite{DBLP:conf/cvpr/PavlakosCGBOTB19} or normalizing flow \cite{DBLP:journals/corr/abs-2003-10350}, we are not using any prior loss to maximize the probability of latent vector in the Gaussian-distributed latent space. This was because such prior loss $E_{prior}$ are encouraging the pose to stay close to a single mean pose, where $z = 0$, which can introduce unnecessary errors. The effect of prior loss $E_{prior}$ will be further explored in the experimental session (Sec.~\ref{sec:ablation}). 
% This also explains why we optimize the vector $z$ in the latent space by calculating $\arg\min_{z} E(f_{decoder}(z))$ rather than directly optimize the 3D pose $P_{seq}$ in the ambient space by calculating $\arg\min_{P_{seq}} E(P_{seq})$. This is because if we optimize $P_{seq},$ we cannot leverage any prior captured by VAE without prior loss $E_{prior}$. However, if we optimize $z$, we can still incorporate the pose prior without prior loss $E_{prior}$ as $P_{seq}$ in $E(P_{seq})$ is obtained from the decoder of the VAE network.

\begin{figure}
	\begin{center}
		\includegraphics[width=0.97\linewidth]{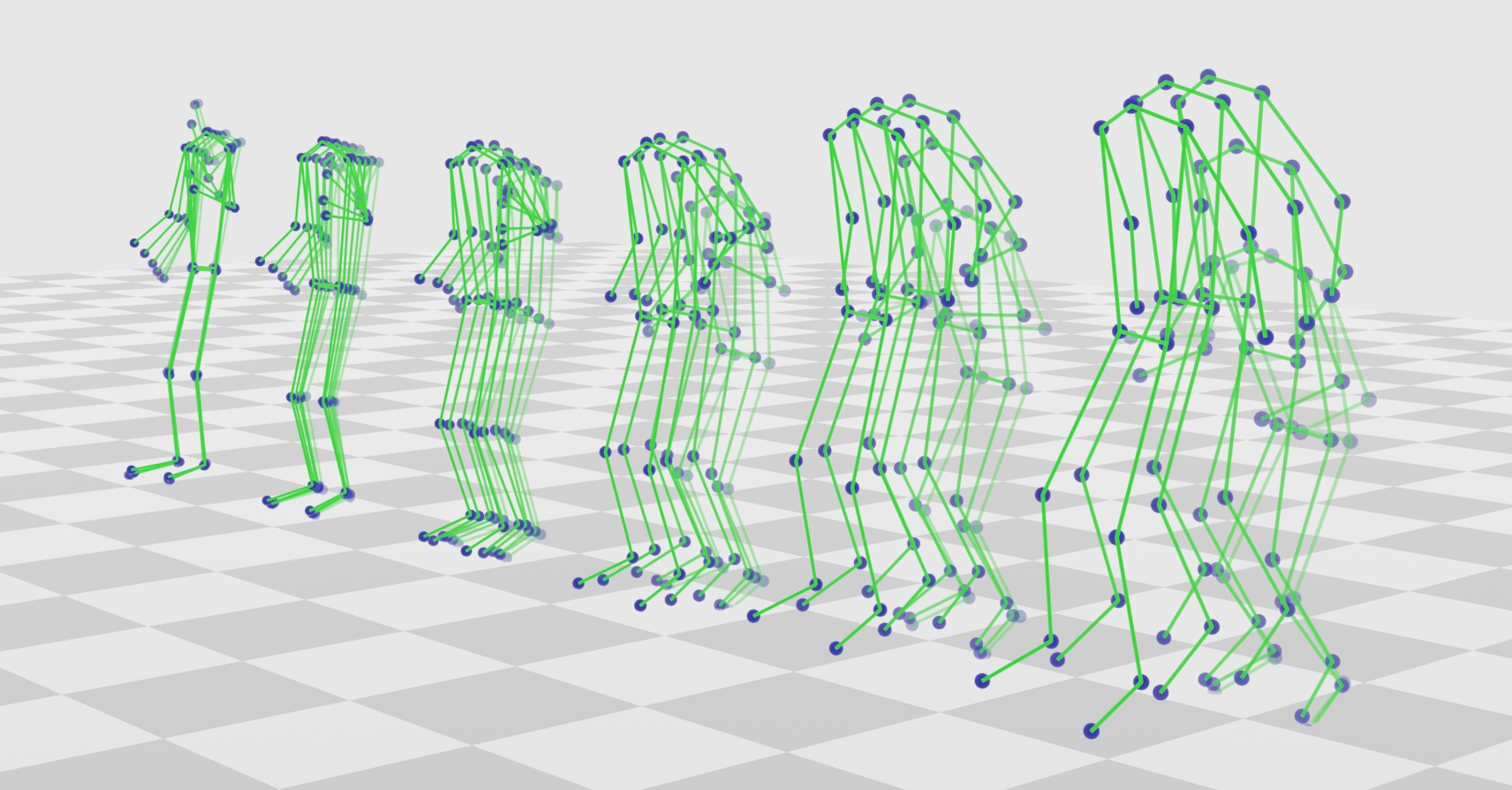}
	\end{center}
	\caption{Interpolation in the latent space. The leftmost and rightmost pose sequences (waving hands and jumping) are reconstructed from two randomly sampled latent vectors, and intermediate pose sequences are reconstructed from linear interpolation between the left and right latent vectors. 
% 	From left to right: interpolation in latent space of sequential VAE between two randomly sampled motions (waving hands and jumping).
	}
	\label{fig:interpolate}
\end{figure}

\subsection{Global Pose Estimation}\label{sec:global_pose}
% \JW{In order to get the global pose, we first transform the optimized local poses from the local fisheye camera space to the world coordinate space with the camera poses estimated by a monocular SLAM method to get initial global poses (Sec.~\ref{subsec:slam}). 
% To improve the initial global poses in terms of accuracy and temporal stability, we perform a spatio-temporal optimization over global poses (Sec.~\ref{subsec:global_optim}). }

Based on the pose optimized by the local pose optimizer, we seek to get the 3D pose in the global coordinate system. We firstly use the monocular SLAM to get the camera pose sequence and project the local pose sequence to the global space (Sec.~\ref{subsec:slam}), then we optimize the initial global pose sequence with our global pose optimizer (Sec.~\ref{subsec:global_optim}).

\subsubsection{Initialization} \label{subsec:slam}
% \subsubsection{SLAM} \label{subsec:slam}
To obtain the initial global body poses, we first estimate the camera poses using ORB-SLAM2 \cite{murTRO2015}. \
% Note that to avoid that the effects caused by the moving person in the egocentric view, we employ a fixed mask to filter out the human body in the view. 
In order to avoid the effects caused by the moving person in the egocentric view, we employ a square-shaped mask that roughly covers a large portion of the body to remove most of the feature points detected on the main body parts. We use a fixed mask rather than estimating a silhouette mask for each image for the sake of effectiveness and robustness.

With the estimated camera pose $(R_i, t_i)$ ($i=1,\cdots,B$), the local body pose $P_{i}$ can be transformed into the world coordinate space to obtain its initial global body pose $\widetilde{P}^g_{i}$: 

\begin{equation}
    \widetilde{\mathcal{P}}^g_{i} = R_i \cdot \mathcal{P}_{i} + t_i,   \widetilde{\mathcal{P}}^g_{i} \in \widetilde{\mathcal{P}}^g_{seq}
\end{equation}
where $\widetilde{\mathcal{P}}^g_{seq}$ is the corresponding inital global pose segment of ${\mathcal{P}}_{seq}$.
% In this section, we use the monocular ORB-SLAM2 \cite{murTRO2015} to get the global pose of the egocentric camera. In order to avoid the influence of human body in the scene, we use semantic segmentation network \cite{WangSCJDZLMTWLX19} which was trained on the Mo$^2$Cap$^2$ synthetic data to segment out the human body. We get the scale of camera trajectory by calibrating the camera position with a checkerboard in the first several frames of the sequence. We choose to use the SLAM to get the camera pose rather than using deep-learning based visual odometry system \cite{DBLP:conf/icra/ZhanWB020,DBLP:conf/cvpr/YangSWC20} because the SLAM is more robust to different scenes. In this way, we can transform each frame $P_{i}$ in the optimized local pose sequence $P_{seq}$ to the frames $P_{i}^g$ of the global pose sequence $P_{seq}^g$ with the corresponding camera pose $R_i, t_i$:
% \begin{equation}
%     P_{i}^g = R_i \cdot P_{i} + t_i
% \end{equation}

\subsubsection{Global Pose Optimizer} \label{subsec:global_optim}
\begin{figure}
	\begin{center}
		\includegraphics[width=0.97\linewidth]{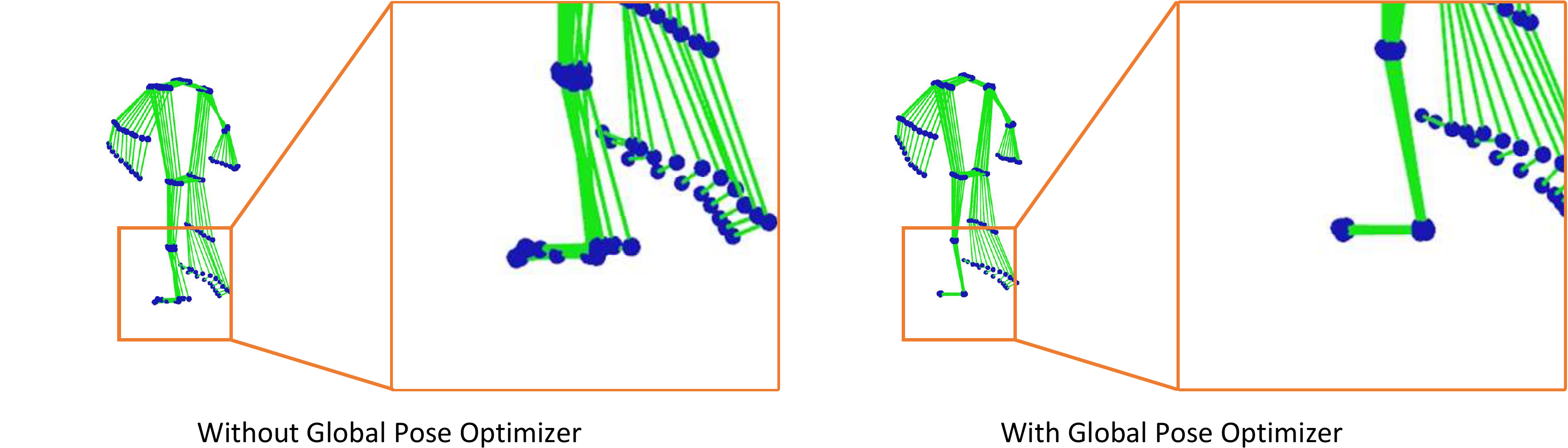}
	\end{center}
	\caption{The global pose with/without global pose optimizer. The left foot is zoomed in for better comparison.
	}
	\label{fig:global_optim}
\end{figure}
% \WX{Can you make the optimization part look simpler: instead of two optimization stages, can you combine the two energy functions in to one with two terms (local pose prior and global pose prior)? Later on, you can say we first optimize without the global pose for X iterations and then optimize with the whole function for Y iterations. A reviewer mentioned there are too many steps. This simpler version might address this problem.}
Simply combining local poses with camera poses would not achieve very high-quality global poses because the optimized local body poses are not constrained to be consistent with the corresponding camera poses. 
For example, the initial global pose in the left part of Fig.~\ref{fig:global_optim} suffers from the footskate artifact, which means the foot moves when it should remain in a fixed position on the ground. 
In order to alleviate such inconsistency errors, we perform another spatio-temporal optimization on the initial global pose.
% To enforce this constraint, we apply another spatio-temporal optimization as described in Sec.~\ref{sec:local_pose}. 
We first train a sequential VAE on global pose sequences from the AMASS dataset in the same way presented in Sec.~\ref{subsec:vae}. To measure the smoothness of our learned latent space, we conducted an experiment of interpolating two different body motions. The results shown in Fig.~\ref{fig:interpolate} demonstrate that the learned latent space is smooth (also see this result in the supplemental video), which is important for the subsequent optimization process.
With the learned latent space of global motion, we seek for a latent vector $z^g$ such that the global pose sequence $\mathcal{P}_{seq}^g = f_{dec}^g(z^g)$ minimizes the following objective function:

% After transforming the local 3D pose to global space with the egocentric camera pose, there can be still mismatch between body motion and global transformation, which will cause artifacts like the sliding feet on the ground. To further tackle this issue, we use similar optimization approach as Sec.~\ref{sec:local_pose} to get more realistic human motion in the global space.

% We firstly train a sequential VAE, which constitutes of 1D convolutional encoder $f_{encoder}^g$ and 1D de-convolutional decoder $f_{decoder}^g$, on the global pose sequence from the AMASS dataset with the method presented in Sec.~\ref{subsec:vae}. The captured global motion prior enable the direct optimization in the global coordinate system. With this motion prior, we mitigate the possible artifacts in the global motion in the following optimization process.

% In the optimization process, we try to find a latent vector $z^g$ such that we can get the global pose sequence $P_{seq}^g = f_{decoder}^g(z^g)$ which minimizes the objective function $E(P^g_{seq})$:
\begin{equation}
\begin{aligned}
E (\mathcal{P}^g_{seq}) &= \lambda_J E_{J}(\mathcal{P}^g_{seq}, \widetilde{\mathcal{P}}^g_{seq}) + \lambda_T E_{T}(\mathcal{P}^g_{seq}) \\
&+ \lambda_B E_{B}(\mathcal{P}^g_{seq})
\end{aligned}
\end{equation}
where $E_{J}(.), E_{T}(.), E_{B}(.)$ are the same as those in \ref{subsec:local_optim}, and $\lambda_J$, $\lambda_T$ and $\lambda_B$ are set as 0.01, 1 and 0.01, respectively.
The example of optimized result is illustrated in the right part of Fig.~\ref{fig:global_optim}, where the footskate artifact is alleviated due to our global optimizer.

%% file: Sections/experiments.tex
\section{Experiments}

\subsection{Datasets}

% To train our sequential VAE, 
Following \cite{DBLP:journals/tvcg/XuCZRFST19} and \cite{DBLP:conf/iccv/TomePAB19}, we train our local egocentric pose estimators on the synthetic dataset from Mo$^2$Cap$^2$.
We use the AMASS dataset \cite{AMASS:ICCV:2019} to train our sequential VAEs. To make the distribution of joint position in the training data consistent with that in the real-world data, we set a virtual fisheye camera attached to the forehead of the human mesh at a distance similar to our capture settings.
% , thus we are able to calculate the local and the global pose in the fisheye coordinate system for training both the VAEs. 

We evaluate our method on both the real-world dataset from Mo$^2$Cap$^2$ \cite{DBLP:journals/tvcg/XuCZRFST19} and a new egocentric dataset. 
% We have to discard the outdoor sequence of Mo$^2$Cap$^2$ because that it lacks features in the surrounding scene, which makes it hard for SLAM to give accurate results. 
Our new real-world dataset was captured using a head-mounted fisheye camera with the similar camera position as Mo$^2$Cap$^2$ \cite{DBLP:journals/tvcg/XuCZRFST19} while the ground truth 3D poses were acquired using a multi-view motion capture system. This dataset contains around 12k frames of 2 actors wearing different clothes and performing 13 types of actions. 
This dataset will be made publicly available and further details of it are shown in the supplementary materials.
%Different from Mo$^2$Cap$^2$, the sequences in our dataset have enough surrounding features for the SLAM.
% All the sequences are split into 100-frame short sequence before evaluation.
% \JW{Compared with the Mo$^2$Cap$^2$ test set and the $x$R-egopose test set, our test set contains more types of actions and more data with global motions. The Mo$^2$Cap$^2$ test set contains 5591 frames (2 actors, 8 types of actions). The $x$R-egopose test set has 10k frames (3 actors, 6 types of actions). }

\subsection{Evaluation Metrics}
We evaluate our method with three different metrics, namely PA-MPJPE, the bone length aligned MPJPE (BA-MPJPE) and the global MPJPE. They all calculate the Mean Per Joint Position Error (MPJPE) but use different ways of alignment to the ground truth.
% We evaluate our method with three different metrics, the PA-MPJPE, the bone length aligned MPJPE (BA-MPJPE) and the global MPJPE. They all calculate the Mean Per Joint Position Error (MPJPE) following Eq.~\ref{eq:mpjpe} but with different ways of alignment to the ground truth.
% \WX{This equation can be removed. People know how to compute mean joint error.}
% \begin{equation}
%     E(P_{seq}, \hat{P}_{seq}) = \frac{1}{B}\frac{1}{N_j}\sum_{i=1}^{B}\sum_{j=1}^{N_j}\left\Vert P_i^j - \hat{P}_i^j \right\Vert_2 \label{eq:mpjpe}
% \end{equation}
% where $P_i^j$ and $\hat{P}_i^j$ are the 3D points of the ground truth and optimized pose at frame $i$ for joint j, out of $B$ frames and $N_j$ joints.
For \textbf{PA-MPJPE}, we rigidly align the estimated pose of each frame to the ground truth pose $P_{seq}$ using $\hat{P}_{seq}$ with Procrustes analysis \cite{kendall1989survey}. 
For \textbf{BA-MPJPE}, we first resize the bone length of each frame in sequences $\hat{P}_{seq}$ and $P_{seq}$ to the bone length of a standard skeleton. Then, we calculate the PA-MPJPE between the two resulting sequences. For \textbf{Global MPJPE}, we globally align all the poses of each batch (100 frames) to the ground truth using Procrustes analysis. 
Each metric has its own focus. The PA-MPJPE measures the accuracy of a single pose while BA-MPJPE eliminates the effects of body scale. The global MPJPE calculates the accuracy of global joint positions, considering the global translation and rotation.

\subsection{Comparison with State-of-the-art Results}

\begin{table}[h]
\begin{center}
\small
\setlength{\tabcolsep}{0.8mm}{
\begin{tabular}{>{\raggedright}p{3.8cm} >{\centering\arraybackslash}p{1.22cm} >{\centering\arraybackslash}p{1.22cm} >{\centering\arraybackslash}p{1.22cm} }
\hlineB{2.5}
Method & Global MPJPE & PA-MPJPE & BA-MPJPE \\
\hline
\multicolumn{3}{l}{\textbf{Mo$^2$Cap$^2$ test dataset}} \\
\hline
Mo$^2$Cap$^2$+SLAM & 117.4 & 80.48 & 61.40 \\
Mo$^2$Cap$^2$+SLAM+Smooth & 113.0 & 76.92 & 58.25 \\
Mo$^2$Cap$^2$+Ours & \textbf{110.5} & \textbf{69.87} & \textbf{52.90} \\
\hline
$x$R-egopose+SLAM & 114.0 & 71.33 & 55.43 \\
$x$R-egopose+SLAM+Smooth & 112.2 & 70.27 & 54.03 \\
$x$R-egopose+Ours & \textbf{110.1} & \textbf{66.74} & \textbf{50.52} \\
\hlineB{2.5}
\multicolumn{3}{l}{\textbf{Our test dataset}} \\
\hline
Mo$^2$Cap$^2$+SLAM & 141.8 & 102.3 & 74.46 \\
Mo$^2$Cap$^2$+SLAM+Smooth & 135.5 & 96.37 & 70.84 \\
Mo$^2$Cap$^2$+Ours & \textbf{119.5} & \textbf{82.06} & \textbf{62.07} \\
\hline
$x$R-egopose+SLAM & 163.4 & 112.0 & 87.20 \\
$x$R-egopose+SLAM+Smooth & 158.1 & 109.6 & 84.70 \\
$x$R-egopose+Ours & \textbf{134.1} & \textbf{84.97} & \textbf{64.31} \\
\hlineB{2.5}
\end{tabular}}
\end{center}
\caption{The experimental results on Mo$^2$Cap$^2$ test dataset \cite{DBLP:journals/tvcg/XuCZRFST19} and our test dataset. 
% \JW{In this table, Mo$^2$Cap$^2$ (or $x$R-egopose) + SLAM is the global pose obtained by projecting the local predictions of Mo$^2$Cap$^2$ (or $x$R-egopose) by the SLAM. Mo$^2$Cap$^2$ (or $x$R-egopose) + SLAM + Smooth is the smoothed global pose of Mo$^2$Cap$^2$ (or $x$R-egopose).}
Mo$^2$Cap$^2$ (or $x$R-egopose) + Ours is the result of our method based on the predictions of Mo$^2$Cap$^2$ (or $x$R-egopose). Our method outperforms previous state-of-the-art Mo$^2$Cap$^2$ \cite{DBLP:journals/tvcg/XuCZRFST19} and $x$R-egopose \cite{DBLP:conf/iccv/TomePAB19} in all of the three metrics.
% \vspace{-0.5em}
}
\label{table:mo2cap2-global}
\end{table}

\begin{figure*}
	\begin{center}
		\includegraphics[width=1\linewidth]{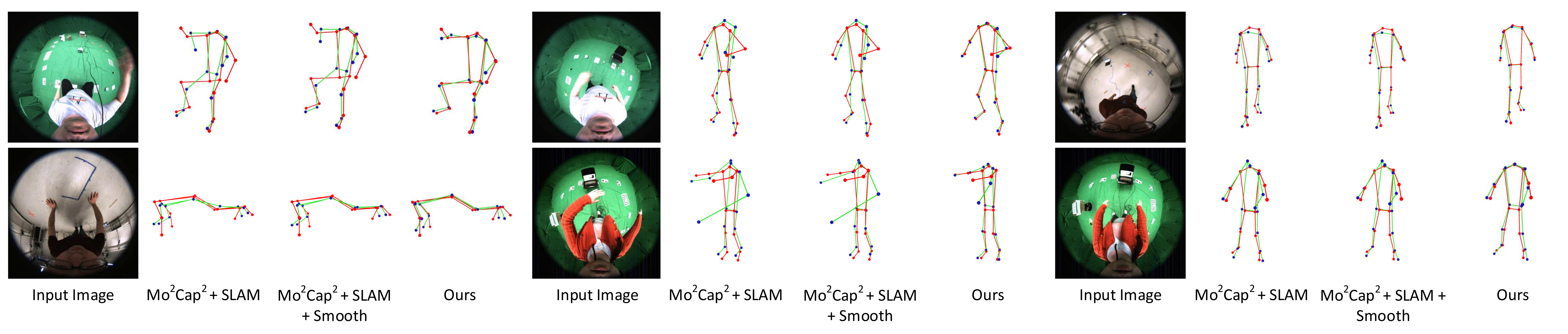}
	\end{center}
	\caption{Qualitative comparison on the accuracy of a single pose. From left to right: input image, Mo$^2$Cap$^2$ result projected with SLAM (green), smoothed Mo$^2$Cap$^2$ result projected with SLAM (green) and our result (green) overlaid on ground truth (red). Note that in order to better show the result, we rigidly align the estimated pose to the ground truth.}
	\label{fig:compare}
\end{figure*}

% \begin{table*}[h]
% \begin{center}
% \small
% \begin{tabular}{l c c c c c c c c c}
% \hlineB{2.5}
% Method & walking & sitting & crawling & crouching & boxing & dancing & stretching & waving & total (mm) \\ \hline
% Mo$^2$Cap$^2$ & 38.41 & 70.94 & 94.31 & 81.90 & 48.55 & 55.19 & 99.34 & 60.92 & 61.40\\
% Mo$^2$Cap$^2$ + Smooth & 37.35 & 64.45 & 87.41 & 69.68 & 45.19 & 54.76 & 90.89 & 49.41 & 58.25\\
% Mo$^2$Cap$^2$ + Ours & \textbf{34.73} & \textbf{56.38} & \textbf{70.19} & \textbf{56.72} & \textbf{39.07} & \textbf{51.71} & \textbf{83.17} & \textbf{41.11} & \textbf{50.20}\\
% \hline
% $x$R-egopose & 39.69 & 63.64 & 64.90 & 61.22 & 47.87 & 58.37 & 84.64 & 53.99 & 55.43 \\
% $x$R-egopose + Smooth & 38.68 & 63.20 & 63.84 & 60.49 & 46.53 & 57.20 & 84.19 & 52.58 & 54.03 \\
% $x$R-egopose + Ours & \textbf{34.45} & \textbf{55.22} & \textbf{61.16} & \textbf{50.85} & \textbf{41.19} & \textbf{51.38} & \textbf{76.91} & \textbf{45.51} & \textbf{48.34}\\
% \hlineB{2.5}
% \end{tabular}
% \end{center}
% \caption{The BA-MPJPE of different types of motions on the indoor sequence of Mo2Cap2 dataset \cite{DBLP:journals/tvcg/XuCZRFST19}. When based on the local poses estimated by Mo$^2$Cap$^2$, our approach improves the Mo$^2$Cap$^2$ \cite{DBLP:journals/tvcg/XuCZRFST19} results by 18.2\% (11.2 mm); when based on the local poses estimated by $x$R-egopose \cite{DBLP:conf/iccv/TomePAB19}, our method improves the $x$R-egopose results by 12.8\% (7.1 mm).}
% \label{table:mo2cap2-BA}
% \end{table*}

Table~\ref{table:mo2cap2-global} compares our approach with previous state-of-the-art single-frame-based methods on our dataset and the indoor sequence of Mo$^2$Cap$^2$ dataset. Since the code or the predictions of $x$R-egopose are not publicly available, we use our implementation instead.
In order to obtain the global pose for Mo$^2$Cap$^2$ and $x$R-egopose, we rigidly transform the local predictions to the world coordinate system with the camera pose estimated by SLAM. This global pose is regarded as our main baseline and denoted as Mo$^2$Cap$^2$ (or $x$R-egopose) + SLAM.
Since the camera poses from ORB-SLAM2 are ambiguous to the scene scale, we further estimate the scale by calibrating the camera position with a checkerboard in the first few frames of the sequence. Note that since the Mo$^2$Cap$^2$ dataset does not provide frames with a checkerboard, we applied the Procrustes analysis to align the trajectory estimated by SLAM with the ground truth trajectory to compute the scale.
For a fair comparison, we also smoothed the global pose of Mo$^2$Cap$^2$ and $x$R-egopose with a Gaussian filter and denote the results as Mo$^2$Cap$^2$ (or $x$R-egopose) + SLAM + smooth.
% To calculate the global MPJPE of the single-frame-based method, we transform the local predictions to the world coordinate system with the camera pose estimated with SLAM.\

\begin{figure*}
	\begin{center}
% 	\includegraphics[width=0.97\linewidth]{Figures/reprojection/reprojection_2_crop.pdf}\\
% 	\vspace{0.3em}
	\includegraphics[width=0.97\linewidth]{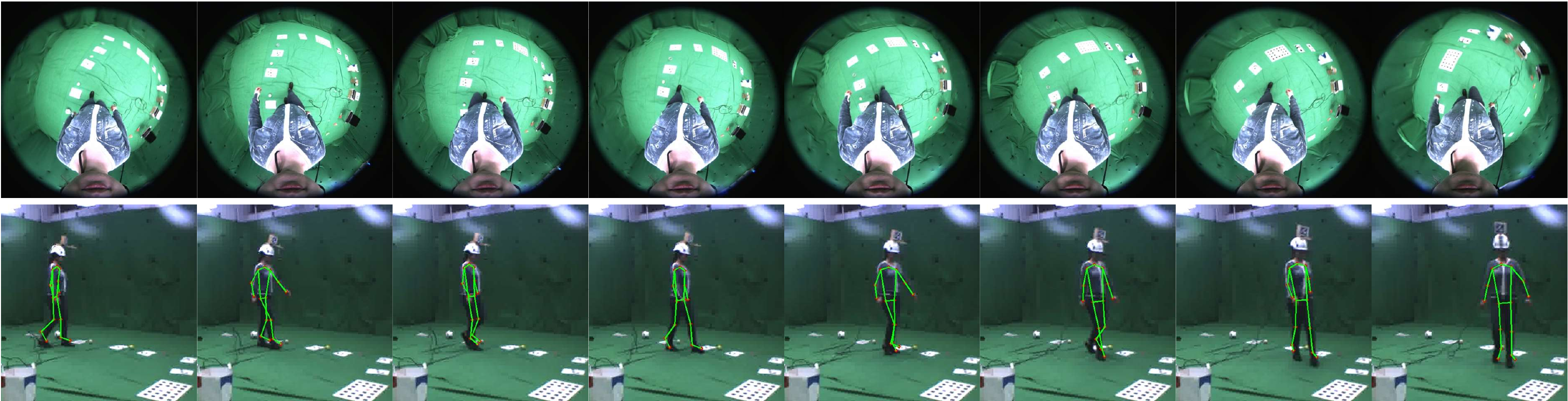}
	\end{center}
	\caption{Global pose estimation results from a third-view camera. Top row: the input egocentric images, bottom row: the estimated 3D pose projected on an external camera.}
	\label{fig:reproj}
% 		\vspace{-1em}
\end{figure*}

From these comparisons, we observe significant improvements, which proves that our method can improve the accuracy of pose estimation results from egocentric videos. 
Please also refer to the supplementary materials for the BA-MPJPE on each type of motion.
% and the BA-MPJPE on each type of motion of our dataset is shown in the Sec.~2 of supplementary materials.
% \WX{More analysis: what actions benefit the most from the global optimization? Does that action that involves locomotion improve the most? What about the actions that have strong occlusion?}
For the qualitative evaluation, we show the comparison between Mo$^2$Cap$^2$ and our method (based on Mo$^2$Cap$^2$) in Fig.~\ref{fig:compare}. Please also see our supplementary video for more results. 
Our method also features the ability to estimate the global body pose, which is shown in Fig.~\ref{fig:reproj} and our supplementary video. In Fig.~\ref{fig:reproj} we demonstrate the accuracy of our global pose estimation by projecting the predicted global pose to an external camera.

\subsection{Ablation Study}\label{sec:ablation}

\begin{table}[h]
\begin{center}
\small
\setlength{\tabcolsep}{1.1 mm}{
\begin{tabular}{l c c c }
\hlineB{2.5}
Method & Global MPJPE & PA-MPJPE & BA-MPJPE \\ \hline
Mo$^2$Cap$^2$ + SLAM & 141.8 & 102.3 & 74.46 \\
\hline
w/o local optim. & 134.7 & 96.33 & 70.77 \\
w/o global optim. & 123.1 & 84.99 & 64.10 \\
\hline
w/o motion prior & 128.1 & 92.31 & 68.10 \\
w. GMM  & 125.0 & 90.12 & 67.50 \\
w. single frame VAE & 122.2 & 87.04 & 65.58 \\
\hline
w. VAE in VIBE & 126.7 & 86.48 & 66.46 \\
w. VAE in MEVA & 121.6 & 84.49 & 63.69 \\
w. MLP based VAE & 122.2 & 85.07 & 65.05 \\
\hline
conventional reproj. & 128.2 & 89.97 & 67.99 \\
% \hline
% $w_{vae}$=1e-3  & 154.0 & 109.9 & 83.91 \\
% $w_{vae}$=5e-4  & 126.0 & 87.76 & 66.64 \\
% $w_{vae}$=1e-4  & 118.0 & 80.97 & 62.43 \\
% $w_{vae}$=1e-5  & 117.4 & 79.47& 61.62 \\
% \hline
% optimize $P_{seq}$ & 124.6 & 90.72 & 68.78 \\
\hline
Mo$^2$Cap$^2$ + Ours & \textbf{119.5} & \textbf{82.06} & \textbf{62.07}\\
\hlineB{2.5}
\end{tabular}}
\end{center}
\caption{The quantitative results of ablation study.}
\label{table:ablation}
 \vspace{-0.8em}
\end{table}

We further conduct experiments to evaluate the effects of individual components of our approach. We use Mo$^2$Cap$^2$ as our local pose estimator for all our ablation studies to make the results comparable.
% For fair comparisons, we all base our methods on the Mo$^2$Cap$^2$ pose estimation module.
% The quantitative results are shown in Table~\ref{table:ablation}.

\paragraph{Local/ global pose optimizer.}

% In our method, firstly we use local pose optimizer to optimize the pose in the camera coordinate system, then we optimize the resulting global sequence with global optimizer.
In this experiment, in order to investigate the efficacy of our local and global optimizer, we evaluate our method after removing the local pose optimizer or the global pose optimizer from our whole pipeline. The results are shown in the 2nd and 3rd row of Table~\ref{table:ablation}, which shows that both of the modules are important to our approach. The heatmap reprojection error in the local pose optimizer ensures that the optimized 3D pose conforms to the constraint of 2D predictions. The VAE prior in the global pose optimizer keeps the movement of body limbs in accordance with the global camera pose, thus improves both on the global MPJPE and the local MPJPEs.

\begin{figure}
	\begin{center}
	\includegraphics[width=0.96\linewidth]{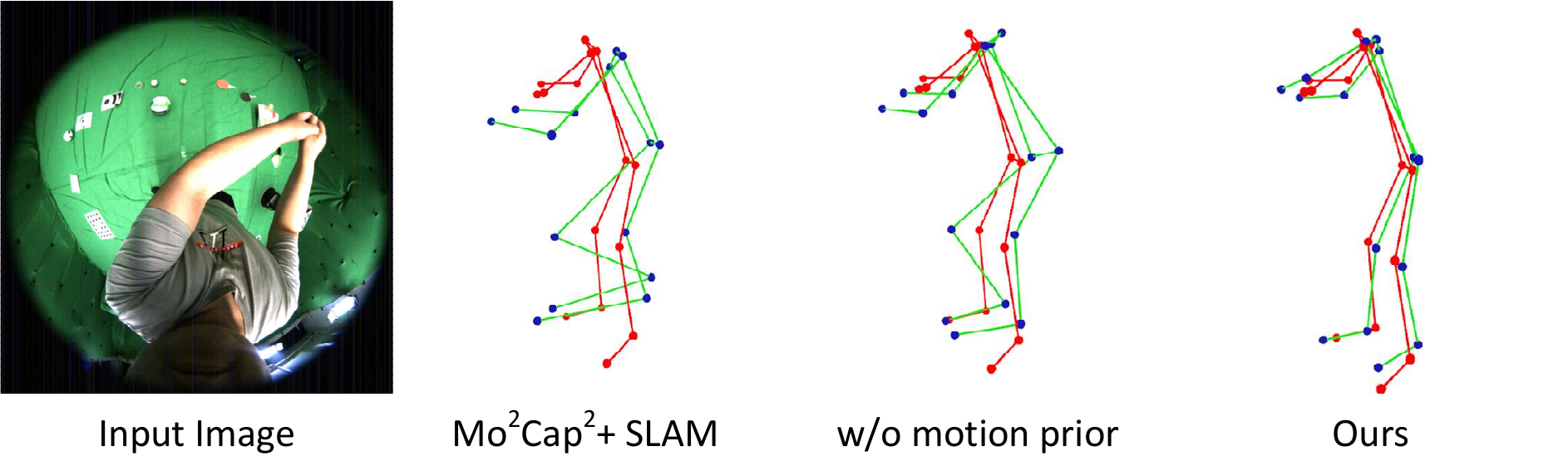}
	\end{center}
	\caption{Comparison between our method with and without motion prior. From left to right: input image, Mo$^2$Cap$^2$ + SLAM (green), the result without motion prior (green) and the one with motion prior (our result) (green) overlaid on the ground truth (red).}
	\label{fig:prior}
\end{figure}

\paragraph{Motion priors.}
In order to validate the importance of motion priors, we test the performance of our optimization framework without our motion priors by directly optimizing 3D pose $P_{seq}$ with $E(P_{seq})$ rather than optimizing the VAE's latent vector $z$. We evaluate the method without motion prior on our dataset and show one of our results in Fig.~\ref{fig:prior}. In this figure, the human leg in the input image is severely occluded. The ambiguity of the image significantly reduced the accuracy of our single-frame pose estimation network. Without the motion prior, our optimization framework cannot resolve the ambiguity and the error is still large, while in our method, the motion prior is able to correct the estimated pose.  The qualitative evaluation in the 4th row of Table~\ref{table:ablation} also confirms our claim. With the motion prior, our spatio-temporal optimization framework is able to make pose predictions smoother and less ambiguous.

We also compared our prior with the gaussian mixture model (GMM) prior used in \cite{DBLP:conf/eccv/BogoKLG0B16, DBLP:conf/cvpr/ArnabDZ19,DBLP:conf/iccv/KolotourosPBD19} and the single-frame VAE prior used in \cite{DBLP:conf/cvpr/PavlakosCGBOTB19}. When comparing with GMM prior, we firstly train the GMM model with 8 Gaussians on the local pose sequence (local GMM) and the global pose sequence (global GMM) from the AMASS dataset. Then we substitute the local and global VAE in our method with the local and global GMM and evaluate three MPJPEs, which is shown in the 5th row of Table~\ref{table:ablation}. GMM prior performs worse since the VAE uses the neural network as a feature extractor, making it easier to capture priors. When comparing with single-frame based VAE prior, we train a VAE network taking a single input pose on the AMASS dataset and substitute the VAE in the local optimizer with the single-frame VAE. The evaluation result is shown in the 6th row of Table~\ref{table:ablation}. The single-frame VAE cannot capture the prior over time, making it less effective than our sequential VAE.
% From the result we can see the motion prior captured by our sequential VAE is the keystone of the success of our method, and it also outperforms the GMM prior and the single-frame VAE prior used in previous works. 

\begin{figure}
	\begin{center}
	\includegraphics[width=0.945\linewidth]{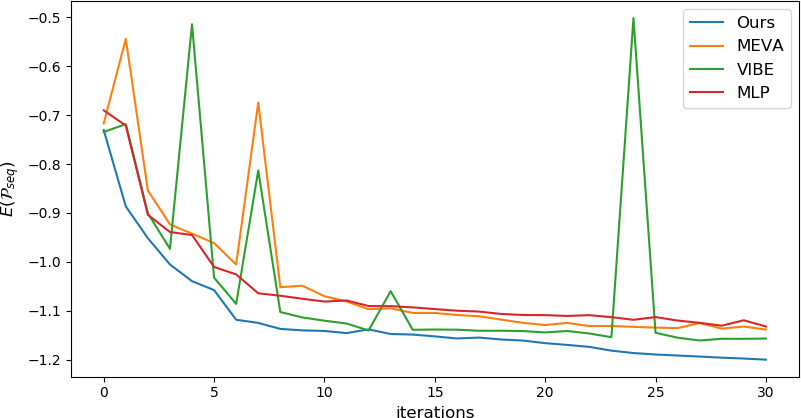}
	\end{center}
	\caption{$E(\mathcal{P}_{seq})$-iteration curve of different VAEs. Our method gives the lowest error while keeping stable during optimization.}
	\label{fig:optimization_vae}
\end{figure}

\paragraph{CNN based sequential VAE.} We use the CNN-based sequential VAE rather than RNN-based VAE for better efficiency and optimization stability. To evaluate our advantage, we substitute our CNN-based sequential VAE in both the local and global optimizer with the VAEs in VIBE~\cite{DBLP:conf/cvpr/KocabasAB20} or MEVA~\cite{Luo_2020_ACCV} (see supplementary materials for implementation details), and report the results in the 7th to 9th rows of Table~\ref{table:ablation}. The result proves that our CNN-based VAE outperforms others in terms of optimization accuracy, which can be attributed to a more stable optimization process. To demonstrate this, we show the the $E(\mathcal{P}_{seq})$-iteration curve of local pose optimization process (Sec.~\ref{subsec:local_optim}) in Fig.~\ref{fig:optimization_vae}, where RNN-based VAEs are less stable due to the gradient explosion issue.
To show the efficiency of CNN-based VAE, we evaluated the time needed for the optimization. Our method takes 195.7ms per 10-frame segment while RNN-based VAE in VIBE and MEVA takes 552.1ms and 1139.4ms per segment respectively. 
We also compared our CNN-based VAE with multilayer perceptron (MLP) based VAE. According to Fig.~\ref{fig:optimization_vae} and the 10th row of Table~\ref{table:ablation}, the MLP-based VAE performs worse since it is not designed to capture the temporal context of the pose sequence.

\begin{figure}
	\begin{center}
	\includegraphics[width=0.96\linewidth]{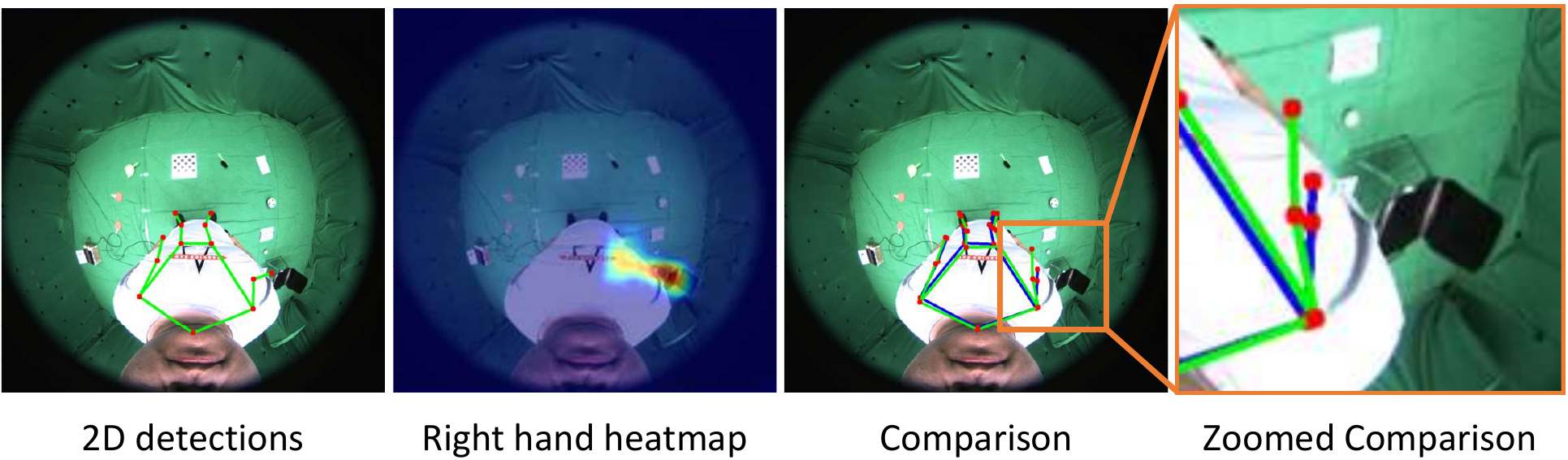}
	\end{center}
	\caption{Comparison between heatmap reprojection error and conventional reprojection error. 
	In the 3rd and 4th image from left, we show the result of heatmap reprojection error in green skeleton and result of conventional reprojection error in blue skeleton.
% 	From left to right: 2D detections, heatmap predictions, the comparison between two reprojection errors and the zoomed comparison result. 
% 	In the third and fourth image, we show result of heatmap reprojection error with green skeleton and result of conventional reprojection error with blue skeleton. 
}
	\label{fig:heatmap}
\end{figure}

\paragraph{Heatmap reprojection error.} In this work we use the heatmap reprojection error while a lot of previous works get the reprojection error by calculating the distance between estimated 2D joints and corresponding projected 3D joints \cite{DBLP:conf/cvpr/ArnabDZ19, DBLP:conf/eccv/BogoKLG0B16, DBLP:conf/cvpr/PavlakosCGBOTB19,DBLP:journals/corr/abs-2003-10350}. To evaluate the improvement of heatmap reprojection error over the previous approach, we substitute the heatmap reprojection error in our pipeline with the conventional reprojection error in \cite{DBLP:conf/eccv/BogoKLG0B16} and compare this with our method. In the qualitative evaluation shown in Fig.~\ref{fig:heatmap}, the 2D pose estimation gives wrong results for the right-hand position while the ground truth hand position is still covered by the heatmap. Our heatmap reprojection error can leverage such uncertainty in the heatmap and gives better results than the conventional reprojection error.  We also show the quantitative result in the 10th row of Table~\ref{table:ablation}. These results validate the advantage of our heatmap reprojection error.
% and its ability to leverage the uncertainty in the predicted heatmap. 

%% file: Sections/conclusions.tex
\section{Conclusions}

In this paper, we propose a method for estimating global poses with a single head-mounted fisheye camera. This is achieved by employing novel strategies in our spatio-temporal optimization framework: (1) a sequential VAE to effectively capture the body motion prior. (2) a global motion prior to ensure consistency between the local body motion and the camera poses. (3) a heatmap-based reprojection error term to leverage the uncertainty in predicted heatmaps. Extensive experiments show that our method outperforms state-of-the-art methods. We further evaluate the effects of individual components of our approach. 

% In this paper, we propose a method for estimating faithful global poses with a single head-mounted fisheye camera.  This is achieved by employing several novel strategies in our spatio-temporal optimization framework: (1) we introduce a sequential VAE to effectively capture the body motion prior. (2) we incorporate global motion prior to ensure consistency between the local body motion and the camera poses estimated by SLAM. (3) we introduce a heatmap-based reprojection error term which leverages the uncertainty in predicted heatmaps to alleviate the adverse influence of the perpective distortion and self-occlusions in the captured video. Extensive experiments show that our method significantly outperforms the state-of-the-art methods. We further evaluate the effects of individual components of our approach. 
% \noindent \textbf{Limitation and future work} As a common limitation for all SLAM methods, our global camera pose estimation requires an environment with rich visual features. 
% Featureless scenes such as white walls lead to unreliable camera poses.

In future work, we will study the solutions to this problem such as the integration of depth sensors. Other future research directions include using the optimized 3D pose in real world to finetune the local pose estimation network and applying our method to the multi-person scenario.

% Other future research directions include using the optimized 3D pose of the captured data in real world to finetune the network for local pose estimation, applying our method to the multi-person scenario, and leveraging other encoding techniques such as transformers to capture better human motion prior. 
% We also aim to visually understand the surrounding scene from the egocentric view, thus enable more applications in the real world environment.